\documentclass{bmvc2k}

\title{Domain Attention Consistency for Multi-Source Domain Adaptation}

\addauthor{Zhongying Deng}{z.deng@surrey.ac.uk}{1, 2}
\addauthor{Kaiyang Zhou}{k.zhou.vision@gmail.com}{3}
\addauthor{Yongxin Yang}{yongxin.yang@surrey.ac.uk}{1}
\addauthor{Tao Xiang}{t.xiang@surrey.ac.uk}{1, 2}

\addinstitution{
 University of Surrey\\
 Guildford, UK
}
\addinstitution{
 iFlyTek-Surrey Joint Research Center on Artificial Intelligence
}
\addinstitution{
 Nanyang Technological University\\
 Singapore
}

\runninghead{Deng, Zhou, Yang, Xiang}{Domain Attention Consistency}

\usepackage{amsmath}
\usepackage{amssymb}
\usepackage{graphicx}
\usepackage{float}
\usepackage{wrapfig}
\usepackage{multirow}
\usepackage{bbm}
\usepackage[title]{appendix}

\makeatletter
\renewcommand\paragraph{\@startsection{paragraph}{4}{\z@}{.1em \@plus1ex \@minus.2ex}{-.5em}{\normalfont\normalsize\bfseries}}
\makeatother

\makeatletter
\newcommand*\bigcdot{\mathpalette\bigcdot@{.5}}
\newcommand*\bigcdot@[2]{\mathbin{\vcenter{\hbox{\scalebox{#2}{$\m@th#1\bullet$}}}}}
\makeatother

\newcommand{\tableCellHeight}{1}
\newcommand{\tabstyle}[1]{
  \setlength{\tabcolsep}{#1}
  \renewcommand{\arraystretch}{\tableCellHeight}
  \centering
}

\newcounter{ablmodel} 
\newcommand{\ablmodel}{\refstepcounter{ablmodel}\theablmodel}

\begin{document}

\maketitle

\begin{abstract}
Most existing multi-source domain adaptation (MSDA) methods minimize the distance between multiple source-target domain pairs via feature distribution alignment, an approach borrowed from the single source setting. However, with diverse source domains, aligning pairwise feature distributions is challenging and could even be counter-productive for MSDA. In this paper, we introduce a novel approach: transferable attribute learning. The motivation is simple: although different domains can have drastically different visual appearances, they contain the same set of classes characterized by the same set of attributes; an MSDA model thus should focus on learning the most transferable attributes for the target domain. Adopting this approach, we propose a domain attention consistency network, dubbed DAC-Net. The key design is a feature channel attention module, which aims to identify transferable features (attributes). Importantly, the attention module is supervised by a consistency loss, which is imposed on the distributions of channel attention weights between source and target domains. Moreover, to facilitate discriminative feature learning on the target data, we combine pseudo-labeling with a class compactness loss to minimize the distance between the target features and the classifier's weight vectors. Extensive experiments on three MSDA benchmarks show that our DAC-Net achieves new state of the art performance on all of them.
\end{abstract}

\section{Introduction}
\label{sec:intro}
The domain shift problem has been one of the main obstacles for large-scale deployment of machine learning systems in real-world applications~\cite{PanY09TKDE,zhou2021domain}. This is because in practice, we often need to apply a trained model to a new target environment where the test data follow a different distribution from the training data. As a result, the performance of the model typically drops significantly. This problem has persisted in the deep learning era, even when deep convolutional neural networks (CNNs) have demonstrated great successes in solving many recognition tasks~\cite{recht2019imagenet,zhou2021domain}. As a key solution to overcome the domain shift problem, unsupervised domain adaptation (UDA) has been extensively studied~\cite{syn_digits,baktashmotlagh2013unsupervised,ganin2016domain,long2016unsupervised,2018Maximum,zhou2020domain,kang2020contrastive}. UDA aims to transfer the knowledge learned from one or multiple labeled source domains to a target domain in which only unlabeled data are given for model adaptation.

\begin{wrapfigure}[14]{r}{0.5\textwidth}
    \centering
    \vspace{-0.3cm}
    \includegraphics[width=0.5\columnwidth]{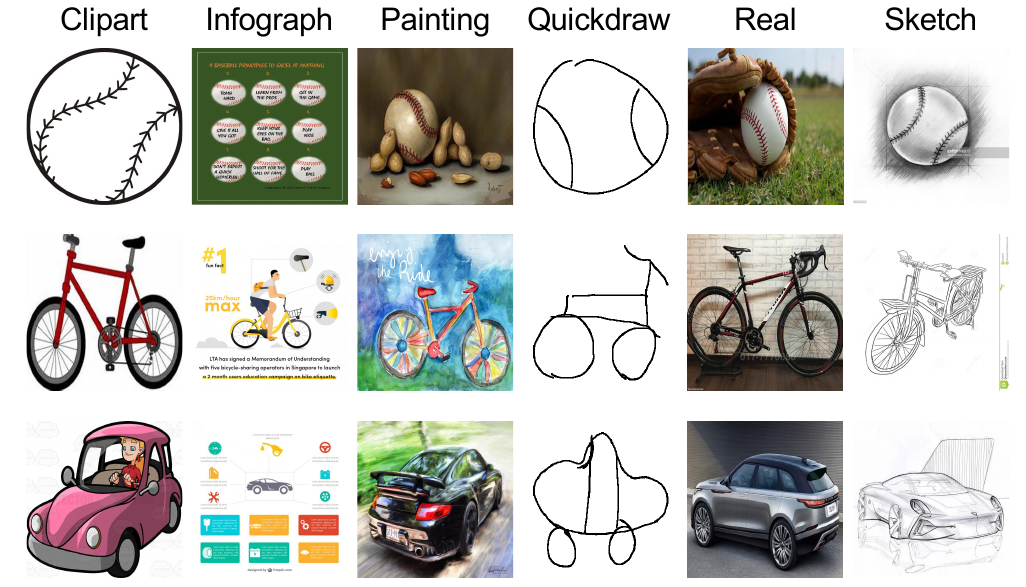}
    \vspace{-0.7cm}
    \caption{Example images from the MSDA benchmark DomainNet~\cite{Peng_2019_ICCV}. Each row contains object images of the same class but from different domains.
    }
    \label{fig:ims_dn}
\end{wrapfigure}

Early UDA work has been focused on the single-source setting~\cite{2018Maximum,lu2020stochastic}. However, in real world, the source training data can often be collected from multiple domains (see Figure~\ref{fig:ims_dn}). This leads to a new UDA setting known as multi-source domain adaptation (MSDA), which has received increasing attention in recent years~\cite{2018Deep,Peng_2019_ICCV,wang2020learning,zhou2020domain}. Most existing MSDA methods are based on aligning the feature distribution of the unlabeled target domain data with those of the source domains.
Feature alignment has been widely used for tackling domain adaptation~\cite{ben2010theory} and is adopted by most single-source UDA methods~\cite{gretton2012kernel,ganin2016domain,tzeng2017adversarial}. However, this approach seems to be much less successful when applied to MSDA. This is not surprising: from Figure~\ref{fig:ims_dn}, it is evident that objects of the same class can have drastically different appearances across different domains. Aligning the feature distribution of a target domain to all the source domains requires a set of features that are completely domain-invariant. This is extremely difficult to achieve; and forcing it can be counter-productive---it has been shown that enforcing such an alignment across multiple source domains can lead to performance  inferior to that of using a single source domain~\cite{Peng_2019_ICCV}.

In this work, we introduce a new approach based on  \emph{transferable attribute learning} as an alternative to the existing feature distribution alignment based approach to MSDA. The motivation is simple: although different domains can have drastically different visual appearances, they contain the same set of classes, which can be explained using the same set of attributes. For example, Figure~\ref{fig:ims_dn} (mid-row) shows that though a bicycle can be depicted in very different ways across the six domains (e.g., in different image styles), it always consists of wheels, frame, seat, handle bar, etc. Some of these attributes are even shared across classes (e.g., cars also have wheels). Therefore, we argue that an MSDA model should focus on learning from source domains the most  transferable attributes for the target domain, which can be achieved by enforcing consistency on attributes used by different domains.

To realize transferable attribute learning, we propose a novel  domain attention consistency network, dubbed DAC-Net. We follow the conventional model design adopted in most papers~\cite{Peng_2019_ICCV,yang2020curriculum}, which consists of a feature embedding network and a classifier (a softmax-activated fully-connected layer) shared by both the source and target domains. However, instead of aligning feature distributions via some distance metrics, we propose to enforce domain attention consistency to identify transferable attributes, each represented by a CNN feature channel. To that end, we first construct a feature channel attention module to encourage the DAC-Net to use a small set of features (latent attributes) to represent each image. Then, to ensure these attributes to be transferable, a novel domain attention consistency loss is introduced, which minimizes the distribution divergence of channel attention weights between each pair of source and target domains. To facilitate discriminate feature learning on the target data, we further combine pseudo-labeling~\cite{lee2013pseudo,sohn2020fixmatch} and a class compactness loss to minimize the distance between the target features and the classifier's weight vectors.

Our contributions are summarized as follows:
{\bf (1)} We propose a new transferable attribute learning based approach to tackle MSDA. The main idea is to learn, in each source domain, the most transferable attributes/features for the target domain.
{\bf (2)} We propose a novel domain attention consistency network (DAC-Net), which aims to align the distributions of channel-wise attention weights in each pair of source-target domains for learning transferable latent attributes.
{\bf (3)} To facilitate discriminative feature learning, we combine pseudo-labeling with a class compactness loss to pull together the target features and the classifier's weight vectors.
{\bf (4)} Extensive experiments on three MSDA datasets, including DomainNet~\cite{Peng_2019_ICCV}, Digit-Five~\cite{1998Gradient,37648,syn_digits} and PACS~\cite{li2017deeper}, show that DAC outperforms the state of the art on all datasets, often by significant margins (e.g., 3.8\% on the largest DomainNet).

\section{Related Work}
\label{sec:related}
\paragraph{Single-source domain adaptation} has been extensively researched. The main stream of domain adaptation methods has been devoted to reducing the distribution mismatch between source and target domains, mostly at the feature level~\cite{gretton2012kernel,long2015learning,balaji2019normalized,ganin2016domain}. 
Direct distance metrics like maximum mean discrepancy (MMD)~\cite{gretton2012kernel} and its kernelized version have been used in~\cite{gretton2012kernel,long2015learning} for distribution divergence minimization. 
Inspired by generative adversarial network (GAN)~\cite{goodfellow2014generative}, adversarial learning has been used to align the feature distributions between source and target domains~\cite{ganin2016domain,tzeng2017adversarial}. 
Recent work has further taken into account the class information, and focused on class-wise feature alignment across domains by using bi-classifiers~\cite{2018Maximum,lu2020stochastic} or aligning class centroids~\cite{xie2018learning,kang2019contrastive}.

More related to our work are attention alignment-based methods~\cite{kang2018deep,wang2019transferable}. In~\cite{kang2018deep}, the spatial attentions summarized across channels between each source domain and the translated pseudo-target domain (via CycleGAN~\cite{zhu2017unpaired}) are aligned. In~\cite{wang2019transferable}, transferable image regions are identified based on adversarial networks. Different from these methods, our design of DAC-Net aims to identify the most transferable attributes (feature channels) by aligning the \emph{distributions of channel attentions} between each pair of source and target domains.

\paragraph{Multi-source domain adaptation (MSDA)} assumes access to multi-source data, compared with the single-source setting. 
Most existing MSDA methods are still based on feature alignment. Xu et al.~\cite{2018Deep} developed deep cocktail network (DCTN), which extends the domain-adversarial learning~\cite{ganin2016domain,tzeng2017adversarial} by learning a domain discriminator for each source-target pair.
Li et al. ~\cite{li2018extracting} chose a relevant subset of each domain to apply feature alignment. Zhu et al.~\cite{zhu2019aligning} proposed to align each source domain's distributions with that of target domain in multiple domain-specific feature spaces.
Peng et al.~\cite{Peng_2019_ICCV} introduced M$^3$SDA, which minimizes moment-based distribution distances between each pair of source-target domains, as well as between each source-source pair.
To facilitate feature alignment, Peng et al.~\cite{peng2020domain2vec} measured domain similarity by using a DOMAIN2VEC model to output vectorial representation for each domain.
Zhou et al. ~\cite{zhou2020domain} leveraged complementary information from multiple domain-specific classifiers to form an ensemble for the target domain.
Pernes et al.~\cite{pernes2020tackling} weighed the importance of each source domain for feature alignment.
Yang et al.~\cite{yang2020curriculum} developed curriculum manager for source selection (CMSS), which aims to learn which source domains are more suitable to be aligned with the target domain. Wang et al.~\cite{wang2020learning} investigated interactions between different domains and developed a knowledge graph-based method called LtC-MSDA to promote information propagation from source domains to the target one.

Our DAC-Net differs from the existing MSDA methods in that no feature distribution alignment is attempted. Instead, we first introduce a channel feature attention module to encourage the learned features to capture a set of domain-transferable latent attributes. Then we design a consistency loss to minimize the divergence between the distributions of channel attention weights of each source-target domain pair. This is a much softer constraint than feature distribution alignment, and it is also much more effective (see Table~\ref{tab:compare_stateoftheart}).

\paragraph{Attention mechanism} was initially introduced to focus on specific words in one language when translating a word in the other language~\cite{bahdanau2015neural}. In computer vision, attention has been used for CNN architecture design.
Hu et al.~\cite{hu2018squeeze} investigated attention from the channel dimension instead of the spatial dimension. They designed a squeeze-and-excitation network (SENet), which introduces a light-weight module that produces channel-wise attention values for a CNN layer. Woo et al.~\cite{Woo_2018_ECCV} further developed convolutional block attention mechanism (CBAM), which combines spatial attention~\cite{Wang_2017_CVPR} with channel attention~\cite{hu2018squeeze}.
In this work, for the first time, we exploit attention for \emph{transferable attribute learning} for domain adaptation---a channel attention network is learned to attend to features that are transferable between source domains and the target domain, supervised by a novel DAC loss.

\section{Methodology}

\subsection{Problem Formulation}
In multi-source domain adaptation (MSDA), we are provided with labeled source data from $K$ different domains, $\{ \mathcal{S}_1, ..., \mathcal{S}_K \}$. The training data from the $k$-th source domain are denoted by $\mathcal{S}_k = \{ (x^{\mathcal{S}_k}_i, y^{\mathcal{S}_k}_i) \}_{i=1}^{N_{\mathcal{S}_k}}$ where $x$ and $y$ denote data (image) and label respectively. We also have access to unlabeled data from the target domain, $\mathcal{T} = \{ x^{\mathcal{T}}_i \}_{i=1}^{N_{\mathcal{T}}}$. In this paper, we focus on image  classification problems and assume a shared label space for the  source and target domains. The goal is to train a classification model leveraging $\{ \mathcal{S}_k \}_{k=1}^K$ and $\mathcal{T}$ so that the model can work well on an unseen test set in the target domain.

\subsection{Domain Attention Consistency}
To address MSDA, we propose a  \emph{domain attention consistency} network (DAC-Net). The motivation behind DAC-Net is to learn to attend to features that are transferable between multiple source domains and the target domain. Each feature (a CNN feature channel) represents a particular attribute. Therefore, in essence we aim to identify attributes that are transferable to the target domain.

\begin{figure*}[t]
    \centering
    \includegraphics[trim=5 20 20 20, clip, width=\textwidth]{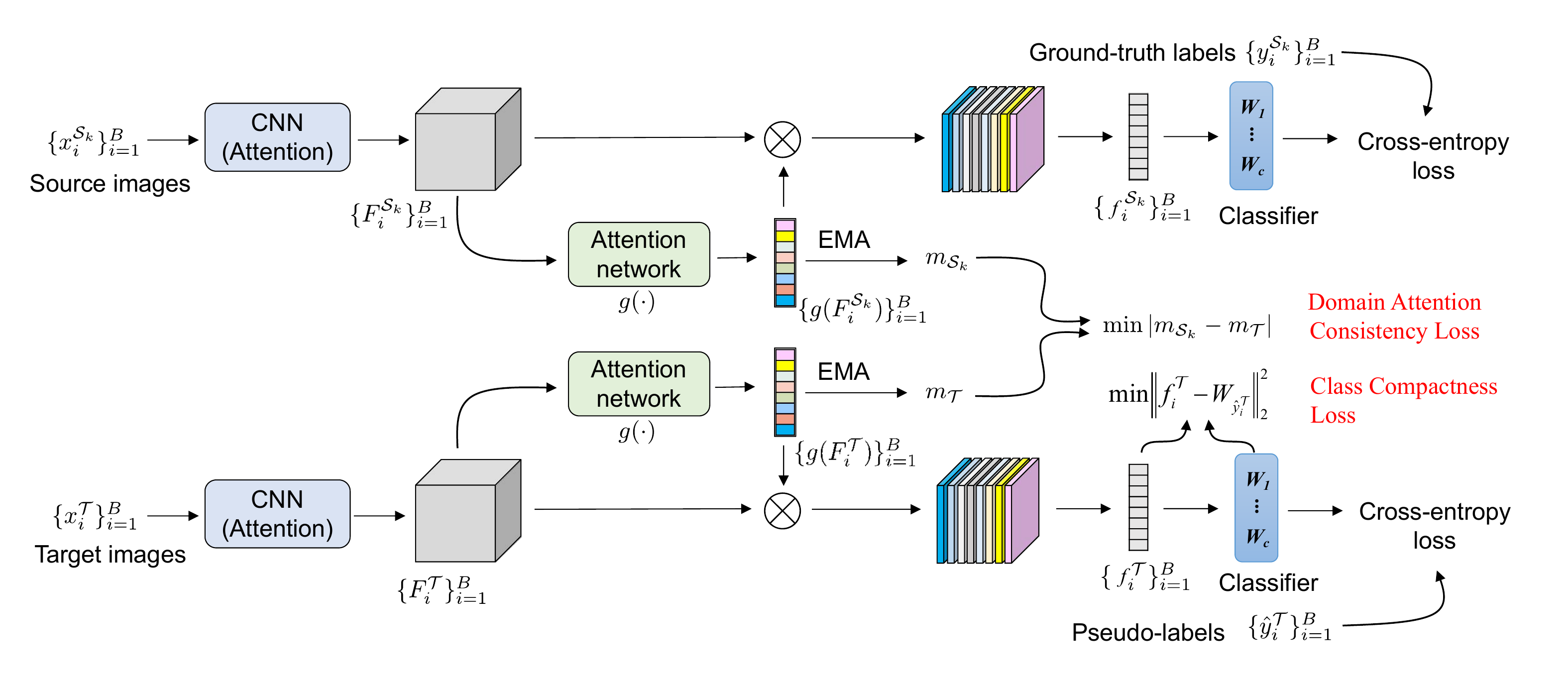}
    \vspace{-0.6cm}
    \caption{Overview of our DAC-Net, designed to  attend to features that are transferable from multiple source domains to the target domain. This is achieved by optimizing a domain attention consistency loss that minimizes the $\ell_{1}$ distance between the exponential moving average (EMA) of attention weights of each source domain ($m_{\mathcal{S}_k}$) and that of the target domain ($m_{\mathcal{T}}$). In implementation, we apply the attention network to multiple layers in a CNN. All the parameters are shared across domains. }
    \label{fig:pipeline}
\end{figure*}

The architecture of our DAC-Net is illustrated in Figure~\ref{fig:pipeline}. It adopts a common architecture design used by existing MSDA models \cite{Peng_2019_ICCV,yang2020curriculum}. Concretely,  the model consists of a feature embedding CNN sub-network followed by a classification layer, both of which are shared across the source and target domains. To encourage the feature embedding CNN to learn a set of features that correspond to transferable latent attributes, we introduce a channel attention module inserted into different layers of the embedding network. With attention modeling for each image, DAC-Net is encouraged to use a subset of the feature channels to explain the image content, therefore facilitating the discovery of transferable latent attributes.

However, without proper supervision, simply inserting attention modules to a CNN network would not help knowledge transfer from source domains to the target (see Table~\ref{tab:ablation_pacs},  \#\ref{ablmodel:Ls_Lt_A} vs.~\#\ref{ablmodel:Ls_Lt}). We therefore propose a novel domain attention consistency (DAC) loss, which minimizes the distance between the \emph{distributions} of attention weights used by source domains and the target domain. Below we detail the design of the attention module and the DAC loss.


\begin{wrapfigure}[7]{r}{0.5\textwidth}
\centering
    \vspace{-0.05cm}
    \includegraphics[trim=0 2 0 0, clip, width=.5\columnwidth]{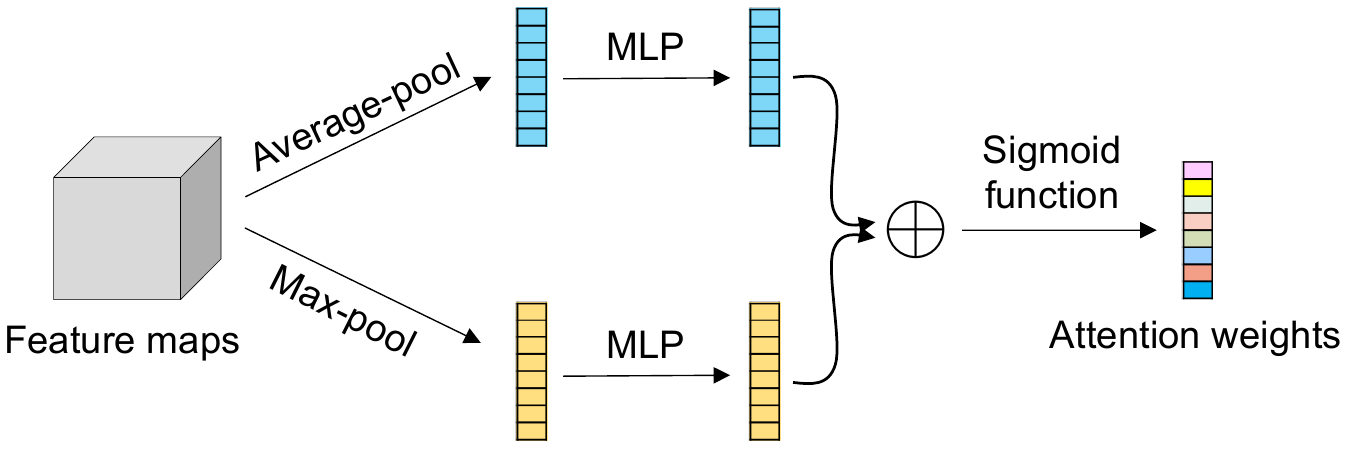}
    \vspace{-0.6cm}
    \caption{Architecture of our attention module}
    \label{fig:attention_network}
\end{wrapfigure}

\paragraph{Attention module.}
Let $F \in \mathbb{R}^{C \times H \times W}$ denote feature maps extracted by a CNN, where $C$, $H$ and $W$ denote channel depth, height and width, respectively.  The attention network $g(\cdot)$ takes as input $F$ and produces a vector of attention weights spanning the channel dimension, $g(F) \in \mathbb{R}^C$. To make $g(\cdot)$ light-weight, we follow the design of CBAM~\cite{Woo_2018_ECCV} when constructing $g(\cdot)$. The architecture is detailed in Figure~\ref{fig:attention_network}. The two branches share the same multi-layer perceptron (MLP), which consists of two fully connected layers with the same dimension for input and output. Notably, the hidden dimension in the MLP is reduced from $C$ to $\frac{C}{r}$, where $r$ is a reduction ratio (fixed to 16), to reduce parameter overhead.

\paragraph{Domain attention consistency loss.}
As shown in Figure~\ref{fig:pipeline}, we first extract feature maps $\{ F_i^{\mathcal{S}_k} \}_{i=1}^B$ and $\{ F_i^{\mathcal{T}} \}_{i=1}^B$ from the $k$-th source domain images $\{ x_i^{\mathcal{S}_k} \}_{i=1}^B$ and the target domain images $\{ x_i^{\mathcal{T}} \}_{i=1}^B$, respectively. Here $B$ denotes the batch size. The feature maps are then forwarded to the attention module $g(\cdot)$. In this work, we simply use the attention weight vector $g(F)$ averaged in each domain to represent the domain-level attention distribution. This average could be easily obtained at each mini-batch, but it will be an inaccurate measure of the domain-wise attribute attention statistics. We thus use  exponential moving average (EMA) over mini-batches. Specifically, the EMA attention weights are computed as
\vspace{-0.15cm}
\begin{align}
m_{\mathcal{S}_k} = \alpha m_{\mathcal{S}_k} + (1 - \alpha) \frac{1}{B} \sum_{i=1}^B g(F_i^{\mathcal{S}_k}),\\
m_{\mathcal{T}} = \alpha m_{\mathcal{T}} + (1 - \alpha) \frac{1}{B} \sum_{i=1}^B g(F_i^{\mathcal{T}}),
\label{eq:m_Sk}
\end{align}
where $\alpha$ is fixed to 0.999. 
Using EMA results in a more accurate estimation of the mean attention weights of the entire population (within each domain), while adding negligible computational overhead. 

The domain attention consistency loss is computed as the $\ell_{1}$ distance between $m_{\mathcal{S}_k}$ and $m_{\mathcal{T}}$. This is done for each pair of source and target domains. Formally, the loss is defined as
\vspace{-0.15cm}
\begin{equation} \label{eq:Ld}
L_d = \frac{1}{K}\sum_{k=1}^K | m_{\mathcal{S}_k} - m_{\mathcal{T}} |.
\end{equation}
We have also tried alternative distance measures such as MMD but found that the EMA-based $\ell_{1}$ distance works better (see Table~\ref{tab:ablation_pacs}, \#\ref{ablmodel:Ls_Lt_A_Ld_MMD} vs.~\#\ref{ablmodel:Ls_Lt_A_Ld}).

\subsection{Discriminative Feature Learning}
The  DAC loss in Eq.~\eqref{eq:Ld} is designed for learning transferable features from multiple source domains. Here we turn to the discriminative feature learning part for classification tasks.

\paragraph{Supervised learning for labeled source data.}
We use the cross-entropy loss to exploit labeled source data for learning discriminative features:
\begin{equation} \label{eq:Ls}
L_s = - \frac{1}{KB} \sum_{k=1}^K \sum_{i=1}^B \log p_{i, y_i^{\mathcal{S}_k}}^{\mathcal{S}_k},
\end{equation}
where $p_{i, y_i^{\mathcal{S}_k}}^{\mathcal{S}_k}$ means the predicted probability on the $y_i^{\mathcal{S}_k}$-th class (the ground truth) for $x_i^{\mathcal{S}_k}$.

\paragraph{Pseudo-labeling for unlabeled target data.}
To overcome the absence of labels for the target data, we resort to pseudo-labeling---a widely used technique in semi-supervised learning (SSL)~\cite{lee2013pseudo,NIPS2019_8749,sohn2020fixmatch}. Specifically, we follow the recently proposed FixMatch~\cite{sohn2020fixmatch} but use the pseudo labels for MSDA rather than SSL. Given a target image, its pseudo-label is obtained by feeding the weakly augmented version of the image to the CNN model and picking the predicted class $\hat{y}_i^{\mathcal{T}}$ that has the maximum probability. A threshold $\tau = 0.95$ is used to filter out low-confidence predictions. The cross-entropy loss is then imposed on the model's output for the strongly augmented version of the image, defined as
\begin{equation} \label{eq:Lt}
L_t = - \frac{1}{B} \sum_{i=1}^B \mathbbm{1}(q(\hat{y}_i^{\mathcal{T}}) \geq \tau) \log p_{i, \hat{y}_i^{\mathcal{T}}}^{\mathcal{T}},
\end{equation}
where $q(\hat{y}_i^{\mathcal{T}})$ is the predicted probability on pseudo-label $\hat{y}_i^{\mathcal{T}}$, and $\mathbbm{1}(\cdot)$ the indicator function.

\paragraph{Enforcing class compactness on unlabeled target data.}
To further promote discriminative feature learning on the target data, we design a class compactness loss to encourage the target features to be close to the corresponding classification weight vectors, which can be seen as class prototypes~\cite{snell2017prototypical}.  Let $W_j$ be the weight vector for class $j$ in the last fully-connected layer, and $f_i^{\mathcal{T}}$ the features of $x_i^{\mathcal{T}}$, the class compactness loss is formulated as
\begin{equation} \label{eq:Lc}
L_c = \frac{1}{B} \sum_{i=1}^B \mathbbm{1}(q(\hat{y}_i^{\mathcal{T}}) \geq \tau) ||f_i^{\mathcal{T}} - W_{\hat{y}_i^{\mathcal{T}}} ||_2^2.
\end{equation}

\subsection{Training}
For training the classification CNN model, we combine the losses in Eqs.~\eqref{eq:Ld},~\eqref{eq:Ls},~\eqref{eq:Lt}  and~\eqref{eq:Lc}:
\begin{equation} \label{eq:final_loss}
L = L_s +  L_t + \lambda_{c} L_c + \lambda_d L_d,
\end{equation}
where $\lambda_{c}$ and $\lambda_d$ are hyper-parameters.
The final CNN model trained with Eq.~\eqref{eq:final_loss} is called domain attention consistency network, or DAC-Net.

\section{Experiments}

\subsection{Experimental Setting}

We apply the attention module to multiple layers in our DAC-Net: on Digit-Five, the attention module is applied after the 2nd and 3rd convolution layers; on PACS and DomainNet, where the ResNet architecture is used, we apply the attention module after the \texttt{conv4\_x} and \texttt{conv5\_x} blocks (i.e.~last two residual blocks). We will evaluate this design choice later. Throughout the experiments, $\lambda_d$ is set to 0.3 and $\lambda_{c}$ is 0.1 (unless otherwise specified). 

More settings such as the datasets, protocols, and other training details can be found in the Supplementary Material. The code will be available at \url{https://github.com/Zhongying-Deng/DAC-Net}. 


\begin{table*}[t]
\centering
\caption{Results on three MSDA benchmark datasets where our DAC-Net achieves state-of-the-art performance on all datasets, with a clear margin over other competitors.}
\subtable[DomainNet.]{
   \resizebox{\textwidth}{!}{
    \tabstyle{2pt}
    \begin{tabular}{l|cccccc|c}
    \hline
        \textbf{Methods} & \textbf{Clipart} & \textbf{Infograph} & \textbf{Painting} & \textbf{Quickdraw} & \textbf{Real}  & \textbf{Sketch} & \textbf{Avg}\\
    \hline \hline
        Source-only~\cite{Peng_2019_ICCV} &47.6$\pm$0.52 &13.0$\pm$0.41 & 38.1$\pm$0.45 &13.3$\pm$0.39 &51.9$\pm$0.85 &33.7$\pm$0.54 &32.9\\
        DANN~\cite{ganin2016domain} &45.5$\pm$0.59 &13.1$\pm$0.72 &37.0$\pm$0.69 &13.2$\pm$0.77 &48.9$\pm$0.65 &31.8$\pm$0.62 &32.6\\
        DCTN~\cite{2018Deep} &48.6$\pm$0.73 &23.5$\pm$0.59 &48.8$\pm$0.63 &7.2$\pm$0.46 &53.5$\pm$0.56 &47.3$\pm$0.47 &38.2 \\
        MCD~\cite{2018Maximum} &54.3$\pm$0.64 &22.1$\pm$0.70 &45.7$\pm$0.63 &7.6$\pm$0.49 &58.4$\pm$0.65 &43.5$\pm$0.57 &38.5\\
        M$^3$SDA~\cite{Peng_2019_ICCV} &58.6$\pm$0.53 &26.0$\pm$0.89 &52.3$\pm$0.55 &6.3$\pm$0.58 &62.7$\pm$0.51 &49.5$\pm$0.76 &42.6\\
        CMSS~\cite{yang2020curriculum} &64.2$\pm$0.18 &28.0$\pm$0.20 &53.6$\pm$0.39 &16.0$\pm$0.12 &63.4$\pm$0.21 &53.8$\pm$0.35 &46.5\\
        LtC-MSDA~\cite{wang2020learning} &63.1$\pm$0.50 &\textbf{28.7}$\pm$0.70 &56.1$\pm$0.50 &16.3$\pm$0.50 &66.1$\pm$0.60 &53.8$\pm$0.60 &47.4\\
    \hline
        DAC-Net (\emph{ours}) &\textbf{72.5}$\pm$0.04	&27.6$\pm$0.10	&\textbf{57.8}$\pm$0.06	  &\textbf{23.0}$\pm$0.14	&\textbf{66.7}$\pm$0.10	&\textbf{59.5}$\pm$0.12	&\textbf{51.2}\\
    \hline
    \end{tabular}
    \label{tab:domainnet}
    }
}
~
\subtable[Digit-Five.]{
    \tabstyle{4pt}
    \begin{tabular}{l|ccccc|c}
    \hline
    \textbf{Methods} & \textbf{MNIST} & \textbf{USPS} & \textbf{MNIST-M} & \textbf{SVHN} & \textbf{Synthetic} & \textbf{Avg}\\
    \hline \hline
    Source-only~\cite{yang2020curriculum}  & 92.3$\pm$0.91 &90.7$\pm$0.54  &63.7$\pm$0.83 &71.5$\pm$0.75 &83.4$\pm$0.79 &80.3\\ 
    DANN~\cite{ganin2016domain} &97.9$\pm$0.83 &93.4$\pm$0.79 &70.8$\pm$0.94 &68.5$\pm$0.85 &87.3$\pm$0.68 &83.6\\ 
    DCTN~\cite{2018Deep} & 96.2$\pm$0.80  &92.8$\pm$0.30 &70.5$\pm$1.20 &77.6$\pm$0.40 &86.8$\pm$0.80 &84.8 \\
    MCD~\cite{2018Maximum}  & 96.2$\pm$0.81 &95.3$\pm$0.74 &72.5$\pm$0.67 &78.8$\pm$0.78 &87.4$\pm$0.65 &86.1\\ 
    M$^3$SDA~\cite{Peng_2019_ICCV}   &  98.4$\pm$0.68 &96.1$\pm$0.81 &72.8$\pm$1.13 &81.3$\pm$0.86 &89.6$\pm$0.56 &87.6\\ 
    CMSS~\cite{yang2020curriculum} &99.0$\pm$0.08 &97.7$\pm$0.13 &75.3$\pm$0.57 &88.4$\pm$0.54 &93.7$\pm$0.21 &90.8 \\ 
    LtC-MSDA~\cite{wang2020learning}   &99.0$\pm$0.40 &98.3$\pm$0.40 &85.6$\pm$0.80 &83.2$\pm$0.60 &93.0$\pm$0.50 &91.8\\
    \hline
    DAC-Net (\emph{ours}) &\textbf{99.2}$\pm$0.03 &\textbf{98.7}$\pm$0.11  &\textbf{86.0}$\pm$0.44  &\textbf{91.6}$\pm$0.16  &\textbf{97.1}$\pm$0.18 & \textbf{94.5} \\ 
    \hline
    \end{tabular}
    \label{tab:digit-five}
}
~
\subtable[PACS.]{
    \tabstyle{10pt}
    \begin{tabular}{l|cccc|c}
    \hline
    \textbf{Methods} & \textbf{ArtPainting} & \textbf{Cartoon} & \textbf{Sketch} & \textbf{Photo} & \textbf{Avg}\\
    \hline \hline
    Source-only &81.22	&78.54	&72.54	&95.45	&81.94\\
    MDAN~\cite{zhao2018adversarial} &83.54 &82.34 &72.42 &92.91 &82.80\\
    DCTN~\cite{2018Deep}  &84.67 &86.72 &71.84 &95.60 &84.71\\
    M$^3$SDA~\cite{Peng_2019_ICCV}  &84.20 &85.68 &74.62 &94.47 &84.74\\
    MDDA~\cite{zhao2020multi} &86.73 &86.24 &77.56 &93.89 &86.11\\
    LtC-MSDA~\cite{wang2020learning} &90.19 &90.47 &81.53 &97.23 &89.85\\
    \hline
    DAC-Net (\emph{ours}) &\textbf{91.39}	&\textbf{91.39}	&\textbf{84.97}	&\textbf{97.93}	&\textbf{91.42} \\
    \hline
    \end{tabular}
    \label{tab:pacs}
}
\label{tab:compare_stateoftheart}
\vspace{-0.37cm}
\end{table*}

\subsection{Main Results}
In this section, we compare our DAC-Net with the current state of the art on three MSDA benchmark datasets, namely DomainNet, Digit-Five and PACS. The results are shown in Table~\ref{tab:compare_stateoftheart}. Below we discuss the results in detail.

\paragraph{DomainNet}
is the most challenging dataset among the three due to its large scale. Among the compared methods, the most related  to ours are those based on the idea of domain alignment, including M$^3$SDA and DCTN. In particular, M$^3$SDA minimizes the moment distance between each pair of source-target domains and each pair of source-source domains, while DCTN applies adversarial learning (similar to that used by DANN) to align the feature distribution in each pair of source-target domains. DAC-Net significantly outperforms both M$^3$SDA and DCTN with a significant margin of more than 8.6\%. This improvement demonstrates that aligning attention weights for identifying transferable features is much more useful than aligning feature distributions for MSDA. Compared with the latest methods, i.e.~CMSS and LtC-MSDA (pseudo-labeling~\cite{zhang2018collaborative} is also adopted in LtC-MSDA when estimating class prototypes for the target domain), DAC-Net is also clearly better---with more than 3.8\% improvement over them. It is noteworthy that the biggest improvements over CMSS and LtC-MSDA are obtained on Quickdraw and Sketch, which are drastically different from the other domains where images are mostly colorized with rich textures (see Figure~\ref{fig:ims_dn}). This suggests that the learned transferable latent attributes are more robust against large domain shift. 

On Infograph domain, our method fails to beat some other methods~\cite{yang2020curriculum,wang2020learning}. This can explained by the existence of irrelevant content in Infograph's images, as shown in Figure~\ref{fig:ims_dn}, which may result in very noisy pseudo labels for discriminative feature learning. To improve the quality of pseudo labels, some regularization methods~\cite{zou2019confidence} can be introduced.

\paragraph{Digit-Five and PACS.}
It is clear that DAC-Net achieves the best performance on all target domains on these two datasets, which further justifies our design of domain attention consistency. The other conclusions drawn above also hold: significant gaps exist between DAC-Net and the most related M$^3$SDA and DCTN; the margins over CMSS and LtC-MSDA are also clear, particularly on those challenging domains (over 3\% improvement) like SVHN in Digit-Five and Sketch in PACS.

\vspace{-0.4cm}
\begin{wraptable}[14]{r}{0.55\textwidth}
    \centering
    \caption{Ablation study on PACS. $\mathcal{A}$: attention network. $\Delta$: accuracy difference versus the source-only baseline.}
    \tabstyle{2pt}
    \begin{tabular}{c|l|cc}
    \hline
    \# & \textbf{Methods}  & \textbf{Avg} & $\Delta$\\
    \hline
    \ablmodel \label{ablmodel:Ls} & $L_s$ 	&81.94 & - \\
    \ablmodel \label{ablmodel:Ls_Lt} &  + $L_t$	&88.86 & +6.92\\
    \ablmodel \label{ablmodel:Ls_Lt_A} &  + $L_t$ + $\mathcal{A}$ &88.34 & +6.40\\
    \ablmodel \label{ablmodel:Ls_Lt_A_Ld} &  + $L_t$ + $\mathcal{A}$ + $L_d$ &90.79 & +8.85\\
    \ablmodel \label{ablmodel:Ls_Lt_A_Ld_Lc} &  + $L_t$ + $\mathcal{A}$ + $L_d$ + $L_c$ (final model) & \textbf{91.42} & \textbf{+9.48}\\
    \ablmodel \label{ablmodel:Ls_Lt_A_Ld_centerloss} &  + $L_t$ + $\mathcal{A}$ + $L_d$ + CenterLoss & 87.52 & +5.58\\
    \ablmodel \label{ablmodel:Ls_Lt_A_Ld_noEMA} &  + $L_t$ + $\mathcal{A}$ + $L_d$ w/o EMA	&88.61 & +6.67\\
    \ablmodel \label{ablmodel:Ls_Lt_A_Ld_MMD} & + $L_t$ + $\mathcal{A}$ + MMD-based $L_d$ &90.23 & +8.29\\
    \hline
    \end{tabular}
    \label{tab:ablation_pacs}
\end{wraptable}

\subsection{Ablation Study}
In this section, we conduct ablation studies on PACS to evaluate the main components in our DAC-Net. Note that all variants are trained using exactly the same training parameters as DAC-Net for fair comparison. The results are reported in Table~\ref{tab:ablation_pacs}. Overall, our final model, DAC-Net, brings the largest improvement of 9.48\% over the source-only baseline.

\paragraph{Significance of $L_d$.}
We first apply the attention network to the pseudo-labeling baseline (\#\ref{ablmodel:Ls_Lt}), and compare the results to see whether the attention network brings any improvement. From the comparison of \#\ref{ablmodel:Ls_Lt} vs.~\#\ref{ablmodel:Ls_Lt_A}, we observe that adding the attention network even brings an adverse effect---the accuracy drops from 88.86\% to 88.34\%. However, by incorporating our domain attention consistency (DAC) loss $L_d$ in model training, the performance is significantly improved from 88.34\% to 90.79\% (\#\ref{ablmodel:Ls_Lt_A} vs.~\#\ref{ablmodel:Ls_Lt_A_Ld}). The results confirm that the DAC is essential for learning transferable features.

\paragraph{Importance of the class compactness loss.}
Our class compactness loss in Eq.~\eqref{eq:Lc} essentially pulls together the target features and the classifier's weight vectors to facilitate discriminative target feature learning. By comparing \#\ref{ablmodel:Ls_Lt_A_Ld_Lc} with \#\ref{ablmodel:Ls_Lt_A_Ld}, we observe that the accuracy is improved by 0.63\% with the class compactness loss. We also compare with the  center loss~\cite{wen2016discriminative}, which enforces class compactness using parameterized class centers.  Similar to our class compactness loss, we discard low-confidence pseudo-labels when updating the parameterized centers for the center loss. By replacing our class compactness loss with the center loss, i.e.~\#\ref{ablmodel:Ls_Lt_A_Ld_Lc} vs.~\#\ref{ablmodel:Ls_Lt_A_Ld_centerloss}, we observe a sharp decrease in accuracy (-3.9\%). This result suggests that parameterized class centers cannot be properly learned, possibly close to the decision boundary due to the noisy target pseudo-label; as such, one should rely more on the classification weight vectors to enforce class compactness---the weight vectors updated/dominated by labeled source data can be less noisy and probably far away from the decision boundary. More  experimental results are provided in the Supplementary Material.

\paragraph{Effectiveness of EMA in $L_d$.}
We train a variant of DAC by removing the EMA part when computing $L_d$ so that the mean attention weights are computed based merely on the current mini-batches, i.e. $m_{\mathcal{D}} = \frac{1}{B} \sum_{i=1}^B g(F_i^{ \mathcal{D}})$.
As a result, the performance drops from 90.79\% (\#\ref{ablmodel:Ls_Lt_A_Ld}) to 88.61\% (\#\ref{ablmodel:Ls_Lt_A_Ld_noEMA}). This result justifies the use of EMA statistics for computing the DAC loss.

\paragraph{Alternative distance function for $L_d$.}
In Eq.~\eqref{eq:Ld}, the distance is measured based on the $\ell_1$ distance between the EMA of domain  attention weights. Here we try an alternative distance function based on maximum mean discrepancy (MMD)~\cite{gretton2012kernel}. Comparing with the EMA version (\#\ref{ablmodel:Ls_Lt_A_Ld}), we observe a decrease in performance for the MMD version (\#\ref{ablmodel:Ls_Lt_A_Ld_MMD}).

\paragraph{Sensitivity of $\lambda_d$ and $\lambda_{c}$.}
Recall that $\lambda_d$ and $\lambda_{c}$ control the weights on $L_d$ (domain attention consistency loss) and $L_c$ (class compactness loss) in Eq.~\eqref{eq:final_loss}, respectively. To evaluate how sensitive the performance is to $\lambda_d$ and $\lambda_{c}$, we first set $\lambda_{c}$ to 0 and linearly increase $\lambda_d$ from 0.1 to 1, which covers a wide value range. The results are shown in Figure~\ref{fig:sensiti_lambda_da}. It can be seen that the accuracy is generally stable with different values for $\lambda_d$ (blue solid line), with the best performance achieved at $\lambda_d = 0.3$. Then we fix $\lambda_d$ to 0.3 and adjust $\lambda_{c}$. The results (red dashed line) indicate that increasing $\lambda_{c}$ seems to result in a (smooth) downward trend in performance, with $\lambda_{c}=0.1$ being the best choice.


\paragraph{Where to apply the DAC loss?}
We evaluate four variants of DAC in the first four rows of Table~\ref{tab:ablation_position_Ld} where domain attention consistency loss is applied after different numbers of residual block at different places. The findings are summarized as follows. 1) Applying the loss after the last two residual blocks gives the best performance. 2) Applying the loss to lower layers worsens the performance. This is expected: we aim to discover transferable latent attributes which are semantic/abstract concepts that only emerge in the top layers of a CNN.

\paragraph{Attention alignment vs.~feature alignment.}
We compare our DAC based approach with the most popular feature distribution alignment (FDA) method in Table~\ref{tab:ablation_position_Ld} (see the last row). With everything else identical including the attention network, for the FDA method, we apply the same domain consistency loss $L_d$ on the final features (the features used by the classifier for classification) to directly align feature distributions. It is clear that FDA results in inferior performance to corresponding DAC variants (see Supplementary Material for more experimental results). This observation supports the main claim of the paper: instead of aligning feature distributions, aligning attribute attention weights is more effective for MSDA.

\begin{minipage}{\textwidth}
   \begin{minipage}{0.43\textwidth}
    \centering
    \makeatletter\def\@captype{figure}\makeatother
    \includegraphics[trim=22 2 15 16, width=\columnwidth]{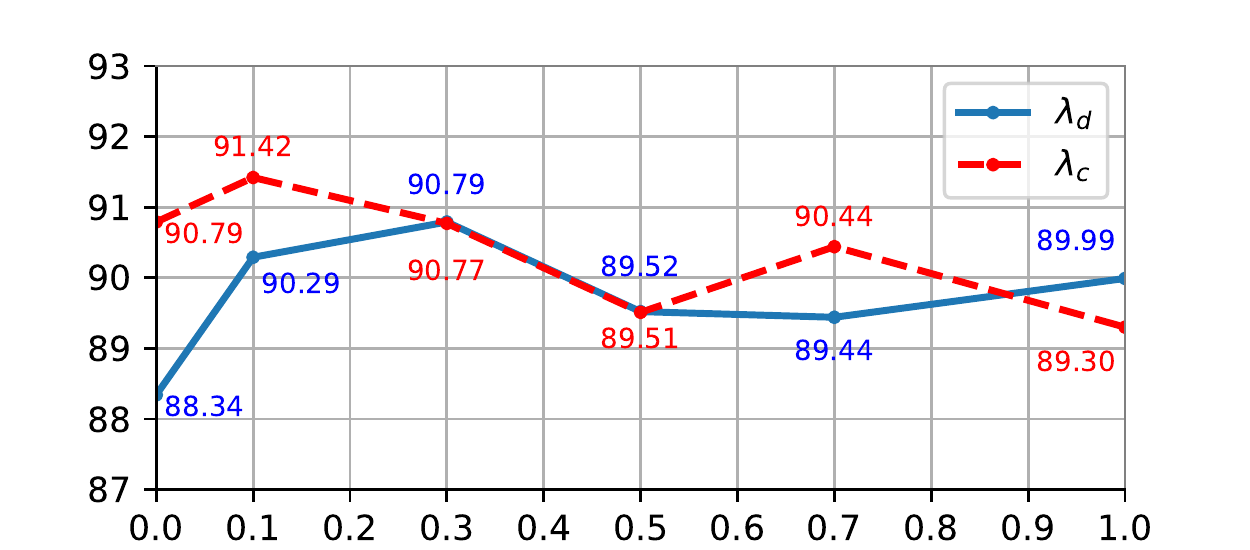}
    \vspace{-0.5cm}
    \caption{Sensitivity of $\lambda_d$ and $\lambda_{c}$ in Eq.~\eqref{eq:final_loss}.}
    \label{fig:sensiti_lambda_da}
  \end{minipage}
  ~
\begin{minipage}{0.5\textwidth}
\vspace{0.2cm}
    \centering
    \makeatletter\def\@captype{table}\makeatother
    \caption{Ablation study on where to apply the DAC loss $L_d$ on PACS. FDA: Feature distribution alignment.}
    \tabstyle{3pt}
    \begin{tabular}{l|l|c}
    \hline
      Methods & Apply $L_d$ after & \textbf{Avg}\\
    \hline
        \multirow{4}{*}{DAC}&Last residual block	&90.48\\
        &Last two residual blocks	&\textbf{90.79}\\
        &Last three residual blocks	&88.46\\
        &All four residual blocks & 88.06\\
    \hline
        FDA &Only final features & 89.27\\
    \hline
    \end{tabular}

    \label{tab:ablation_position_Ld}
    \vspace{-0.2cm}

\end{minipage}
\end{minipage}

\subsection{Visualization}\label{sec:attend_feat}

\begin{figure*}[t]
    \centering
    \includegraphics[width=0.7\textwidth]{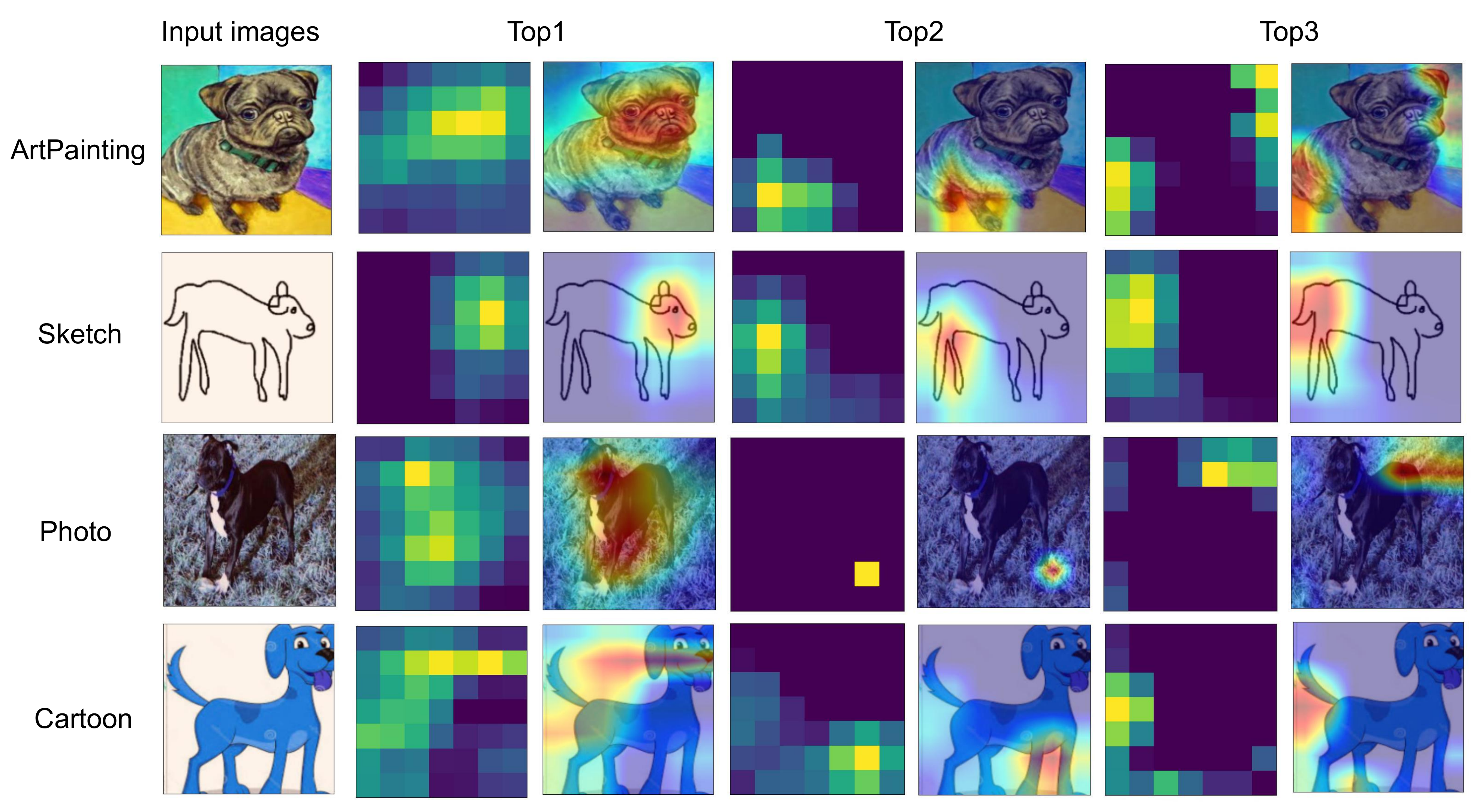}
    \vspace{-0.3cm}
    \caption{Attended feature maps of DAC-Net and their masked input images. We show the feature maps corresponding to the top-3 attention weights across a random testing subset of PACS, and project the feature maps to the input images. These feature maps or masked images are from the same class of dog but different domains.}
    \vspace{-0.2cm}
    \label{fig:vis_feat_map}
\end{figure*}

In this section, we provide visualization of attended feature maps to help understand why our DAC-Net works.
We visualize examples of the top-3 attended feature maps and their corresponding masked images in Figure~\ref{fig:vis_feat_map}.  We can see that the  feature maps with high attentions focus on some semantic attributes, e.g.~the head or leg of a dog, even though their appearance varies greatly across domains.
These semantic attributes are discriminative features for classification, and more importantly are transferable across domains. The ability to discovering them thus underpins the good performance of our DAC-Net.

Visualizations on feature distributions can be found in the Supplementary Material.

\section{Conclusion}
In this paper, we introduced a novel  DAC-Net  to learn transferable latent attributes for MSDA. It incorporates an attention module and a domain attention consistency loss applied on the exponential moving average (EMA) of the attention weights of each source domain and that of the target domain. We also proposed a class compactness loss to pull together the target features and the classification weight vectors (class prototypes).
Extensive experiments on three MSDA benchmark datasets demonstrated that our DAC-Net significantly outperforms the current state-of-the-art competitors.

\bibliography{egbib}

\begin{thebibliography}{53}
\providecommand{\natexlab}[1]{#1}
\providecommand{\url}[1]{\texttt{#1}}
\expandafter\ifx\csname urlstyle\endcsname\relax
  \providecommand{\doi}[1]{doi: #1}\else
  \providecommand{\doi}{doi: \begingroup \urlstyle{rm}\Url}\fi

\bibitem[Bahdanau et~al.(2015)Bahdanau, Cho, and Bengio]{bahdanau2015neural}
Dzmitry Bahdanau, Kyunghyun Cho, and Yoshua Bengio.
\newblock Neural machine translation by jointly learning to align and
  translate.
\newblock In \emph{ICLR}, 2015.

\bibitem[Baktashmotlagh et~al.(2013)Baktashmotlagh, Harandi, Lovell, and
  Salzmann]{baktashmotlagh2013unsupervised}
Mahsa Baktashmotlagh, Mehrtash~T Harandi, Brian~C Lovell, and Mathieu Salzmann.
\newblock Unsupervised domain adaptation by domain invariant projection.
\newblock In \emph{ICCV}, 2013.

\bibitem[Balaji et~al.(2019)Balaji, Chellappa, and Feizi]{balaji2019normalized}
Yogesh Balaji, Rama Chellappa, and Soheil Feizi.
\newblock Normalized wasserstein for mixture distributions with applications in
  adversarial learning and domain adaptation.
\newblock In \emph{ICCV}, 2019.

\bibitem[Ben-David et~al.(2010)Ben-David, Blitzer, Crammer, Kulesza, Pereira,
  and Vaughan]{ben2010theory}
Shai Ben-David, John Blitzer, Koby Crammer, Alex Kulesza, Fernando Pereira, and
  Jennifer~Wortman Vaughan.
\newblock A theory of learning from different domains.
\newblock \emph{ML}, 2010.

\bibitem[Berthelot et~al.(2019)Berthelot, Carlini, Goodfellow, Papernot,
  Oliver, and Raffel]{NIPS2019_8749}
David Berthelot, Nicholas Carlini, Ian Goodfellow, Nicolas Papernot, Avital
  Oliver, and Colin~A Raffel.
\newblock Mixmatch: A holistic approach to semi-supervised learning.
\newblock In \emph{NeurIPS}. 2019.

\bibitem[Chen et~al.(2019)Chen, Chen, Jiang, and Jin]{chen2019joint}
Chao Chen, Zhihong Chen, Boyuan Jiang, and Xinyu Jin.
\newblock Joint domain alignment and discriminative feature learning for
  unsupervised deep domain adaptation.
\newblock In \emph{AAAI}, 2019.

\bibitem[Chen et~al.(2020)Chen, Fu, Chen, Jin, Cheng, Jin, and
  Hua]{chen2020homm}
Chao Chen, Zhihang Fu, Zhihong Chen, Sheng Jin, Zhaowei Cheng, Xinyu Jin, and
  Xian-Sheng Hua.
\newblock Homm: Higher-order moment matching for unsupervised domain
  adaptation.
\newblock In \emph{AAAI}, 2020.

\bibitem[Ganin and Lempitsky(2015)]{syn_digits}
Yaroslav Ganin and Victor Lempitsky.
\newblock Unsupervised domain adaptation by backpropagation.
\newblock In \emph{ICML}, 2015.

\bibitem[Ganin et~al.(2016)Ganin, Ustinova, Ajakan, Germain, Larochelle,
  Laviolette, Marchand, and Lempitsky]{ganin2016domain}
Yaroslav Ganin, Evgeniya Ustinova, Hana Ajakan, Pascal Germain, Hugo
  Larochelle, Fran{\c{c}}ois Laviolette, Mario Marchand, and Victor Lempitsky.
\newblock Domain-adversarial training of neural networks.
\newblock \emph{JMLR}, 2016.

\bibitem[Goodfellow et~al.(2014)Goodfellow, Pouget-Abadie, Mirza, Xu,
  Warde-Farley, Ozair, Courville, and Bengio]{goodfellow2014generative}
Ian Goodfellow, Jean Pouget-Abadie, Mehdi Mirza, Bing Xu, David Warde-Farley,
  Sherjil Ozair, Aaron Courville, and Yoshua Bengio.
\newblock Generative adversarial nets.
\newblock In \emph{NeurIPS}, 2014.

\bibitem[Gretton et~al.(2012)Gretton, Borgwardt, Rasch, Sch{\"o}lkopf, and
  Smola]{gretton2012kernel}
Arthur Gretton, Karsten~M Borgwardt, Malte~J Rasch, Bernhard Sch{\"o}lkopf, and
  Alexander Smola.
\newblock A kernel two-sample test.
\newblock \emph{JMLR}, 2012.

\bibitem[{He} et~al.(2016){He}, {Zhang}, {Ren}, and {Sun}]{7780459}
K.~{He}, X.~{Zhang}, S.~{Ren}, and J.~{Sun}.
\newblock Deep residual learning for image recognition.
\newblock In \emph{CVPR}, 2016.

\bibitem[Hu et~al.(2018)Hu, Shen, and Sun]{hu2018squeeze}
Jie Hu, Li~Shen, and Gang Sun.
\newblock Squeeze-and-excitation networks.
\newblock In \emph{CVPR}, 2018.

\bibitem[Kang et~al.(2018)Kang, Zheng, Yan, and Yang]{kang2018deep}
Guoliang Kang, Liang Zheng, Yan Yan, and Yi~Yang.
\newblock Deep adversarial attention alignment for unsupervised domain
  adaptation: the benefit of target expectation maximization.
\newblock In \emph{ECCV}, 2018.

\bibitem[Kang et~al.(2019)Kang, Jiang, Yang, and
  Hauptmann]{kang2019contrastive}
Guoliang Kang, Lu~Jiang, Yi~Yang, and Alexander~G Hauptmann.
\newblock Contrastive adaptation network for unsupervised domain adaptation.
\newblock In \emph{CVPR}, 2019.

\bibitem[Kang et~al.(2020)Kang, Jiang, Wei, Yang, and
  Hauptmann]{kang2020contrastive}
Guoliang Kang, Lu~Jiang, Yunchao Wei, Yi~Yang, and Alexander~G Hauptmann.
\newblock Contrastive adaptation network for single-and multi-source domain
  adaptation.
\newblock \emph{TPAMI}, 2020.

\bibitem[Kingma and Ba(2014)]{kingma2014adam}
Diederik~P Kingma and Jimmy Ba.
\newblock Adam: A method for stochastic optimization.
\newblock \emph{arXiv preprint arXiv:1412.6980}, 2014.

\bibitem[Lecun and Bottou(1998)]{1998Gradient}
Y~Lecun and L~Bottou.
\newblock Gradient-based learning applied to document recognition.
\newblock \emph{IEEE}, 1998.

\bibitem[Lee(2013)]{lee2013pseudo}
Dong-Hyun Lee.
\newblock Pseudo-label: The simple and efficient semi-supervised learning
  method for deep neural networks.
\newblock In \emph{ICML Workshop}, 2013.

\bibitem[Li et~al.(2017)Li, Yang, Song, and Hospedales]{li2017deeper}
Da~Li, Yongxin Yang, Yi-Zhe Song, and Timothy~M Hospedales.
\newblock Deeper, broader and artier domain generalization.
\newblock In \emph{CVPR}, 2017.

\bibitem[Li et~al.(2018)Li, Murias, Major, Dawson, and
  Carlson]{li2018extracting}
Yitong Li, Michael Murias, Samantha Major, Geraldine Dawson, and David~E
  Carlson.
\newblock Extracting relationships by multi-domain matching.
\newblock In \emph{NeurIPS}, 2018.

\bibitem[Long et~al.(2015)Long, Cao, Wang, and Jordan]{long2015learning}
Mingsheng Long, Yue Cao, Jianmin Wang, and Michael Jordan.
\newblock Learning transferable features with deep adaptation networks.
\newblock In \emph{ICML}, 2015.

\bibitem[Long et~al.(2016)Long, Zhu, Wang, and Jordan]{long2016unsupervised}
Mingsheng Long, Han Zhu, Jianmin Wang, and Michael~I Jordan.
\newblock Unsupervised domain adaptation with residual transfer networks.
\newblock In \emph{NeurIPS}, 2016.

\bibitem[Loshchilov and Hutter(2016)]{loshchilov2016sgdr}
Ilya Loshchilov and Frank Hutter.
\newblock Sgdr: Stochastic gradient descent with warm restarts.
\newblock \emph{arXiv preprint arXiv:1608.03983}, 2016.

\bibitem[Lu et~al.(2020)Lu, Yang, Zhu, Liu, Song, and Xiang]{lu2020stochastic}
Zhihe Lu, Yongxin Yang, Xiatian Zhu, Cong Liu, Yi-Zhe Song, and Tao Xiang.
\newblock Stochastic classifiers for unsupervised domain adaptation.
\newblock In \emph{CVPR}, 2020.

\bibitem[Maaten and Hinton(2008)]{maaten2008visualizing}
Laurens van~der Maaten and Geoffrey Hinton.
\newblock Visualizing data using t-sne.
\newblock \emph{JMLR}, 2008.

\bibitem[Netzer et~al.(2011)Netzer, Wang, Coates, Bissacco, Wu, and Ng]{37648}
Yuval Netzer, Tao Wang, Adam Coates, Alessandro Bissacco, Bo~Wu, and Andrew~Y.
  Ng.
\newblock Reading digits in natural images with unsupervised feature learning.
\newblock In \emph{NeurIPS-W}, 2011.

\bibitem[Pan and Yang(2010)]{PanY09TKDE}
S.J. Pan and Q.~Yang.
\newblock A survey on transfer learning.
\newblock \emph{TKDE}, 2010.

\bibitem[Peng et~al.(2019)Peng, Bai, Xia, Huang, Saenko, and
  Wang]{Peng_2019_ICCV}
Xingchao Peng, Qinxun Bai, Xide Xia, Zijun Huang, Kate Saenko, and Bo~Wang.
\newblock Moment matching for multi-source domain adaptation.
\newblock In \emph{ICCV}, 2019.

\bibitem[Peng et~al.(2020)Peng, Li, and Saenko]{peng2020domain2vec}
Xingchao Peng, Yichen Li, and Kate Saenko.
\newblock Domain2vec: Domain embedding for unsupervised domain adaptation
  supplementary material.
\newblock In \emph{ECCV}, 2020.

\bibitem[Pernes and Cardoso(2020)]{pernes2020tackling}
Diogo Pernes and Jaime~S Cardoso.
\newblock Tackling unsupervised multi-source domain adaptation with optimism
  and consistency.
\newblock \emph{arXiv preprint arXiv:2009.13939}, 2020.

\bibitem[Recht et~al.(2019)Recht, Roelofs, Schmidt, and
  Shankar]{recht2019imagenet}
Benjamin Recht, Rebecca Roelofs, Ludwig Schmidt, and Vaishaal Shankar.
\newblock Do imagenet classifiers generalize to imagenet?
\newblock In \emph{ICML}, 2019.

\bibitem[Saito et~al.(2018)Saito, Watanabe, Ushiku, and Harada]{2018Maximum}
Kuniaki Saito, Kohei Watanabe, Yoshitaka Ushiku, and Tatsuya Harada.
\newblock Maximum classifier discrepancy for unsupervised domain adaptation.
\newblock In \emph{CVPR}, 2018.

\bibitem[Snell et~al.(2017)Snell, Swersky, and Zemel]{snell2017prototypical}
Jake Snell, Kevin Swersky, and Richard~S Zemel.
\newblock Prototypical networks for few-shot learning.
\newblock In \emph{NeurIPS}, 2017.

\bibitem[Sohn et~al.(2020)Sohn, Berthelot, Li, Zhang, Carlini, Cubuk, Kurakin,
  Zhang, and Raffel]{sohn2020fixmatch}
Kihyuk Sohn, David Berthelot, Chun-Liang Li, Zizhao Zhang, Nicholas Carlini,
  Ekin~D. Cubuk, Alex Kurakin, Han Zhang, and Colin Raffel.
\newblock Fixmatch: Simplifying semi-supervised learning with consistency and
  confidence.
\newblock In \emph{NeurIPS}, 2020.

\bibitem[Tzeng et~al.(2017)Tzeng, Hoffman, Saenko, and
  Darrell]{tzeng2017adversarial}
Eric Tzeng, Judy Hoffman, Kate Saenko, and Trevor Darrell.
\newblock Adversarial discriminative domain adaptation.
\newblock In \emph{CVPR}, 2017.

\bibitem[Wang et~al.(2017)Wang, Jiang, Qian, Yang, Li, Zhang, Wang, and
  Tang]{Wang_2017_CVPR}
Fei Wang, Mengqing Jiang, Chen Qian, Shuo Yang, Cheng Li, Honggang Zhang,
  Xiaogang Wang, and Xiaoou Tang.
\newblock Residual attention network for image classification.
\newblock In \emph{CVPR}, 2017.

\bibitem[Wang et~al.(2020)Wang, Xu, Ni, and Zhang]{wang2020learning}
Hang Wang, Minghao Xu, Bingbing Ni, and Wenjun Zhang.
\newblock Learning to combine: Knowledge aggregation for multi-source domain
  adaptation.
\newblock In \emph{ECCV}, 2020.

\bibitem[Wang et~al.(2019)Wang, Li, Ye, Long, and Wang]{wang2019transferable}
Ximei Wang, Liang Li, Weirui Ye, Mingsheng Long, and Jianmin Wang.
\newblock Transferable attention for domain adaptation.
\newblock In \emph{AAAI}, 2019.

\bibitem[Wen et~al.(2016)Wen, Zhang, Li, and Qiao]{wen2016discriminative}
Yandong Wen, Kaipeng Zhang, Zhifeng Li, and Yu~Qiao.
\newblock A discriminative feature learning approach for deep face recognition.
\newblock In \emph{ECCV}, 2016.

\bibitem[Woo et~al.(2018)Woo, Park, Lee, and So~Kweon]{Woo_2018_ECCV}
Sanghyun Woo, Jongchan Park, Joon-Young Lee, and In~So~Kweon.
\newblock Cbam: Convolutional block attention module.
\newblock In \emph{ECCV}, 2018.

\bibitem[Xiao and Zhang(2021)]{xiao2021dynamic}
Ni~Xiao and Lei Zhang.
\newblock Dynamic weighted learning for unsupervised domain adaptation.
\newblock In \emph{CVPR}, 2021.

\bibitem[Xie et~al.(2018)Xie, Zheng, Chen, and Chen]{xie2018learning}
Shaoan Xie, Zibin Zheng, Liang Chen, and Chuan Chen.
\newblock Learning semantic representations for unsupervised domain adaptation.
\newblock In \emph{ICML}, 2018.

\bibitem[Xu et~al.(2018)Xu, Chen, Zuo, Yan, and Lin]{2018Deep}
Ruijia Xu, Ziliang Chen, Wangmeng Zuo, Junjie Yan, and Liang Lin.
\newblock Deep cocktail network: Multi-source unsupervised domain adaptation
  with category shift.
\newblock In \emph{CVPR}, 2018.

\bibitem[Yang et~al.(2020)Yang, Balaji, Lim, and
  Shrivastava]{yang2020curriculum}
Luyu Yang, Yogesh Balaji, Ser-Nam Lim, and Abhinav Shrivastava.
\newblock Curriculum manager for source selection in multi-source domain
  adaptation.
\newblock In \emph{ECCV}, 2020.

\bibitem[Zhang et~al.(2018)Zhang, Ouyang, Li, and Xu]{zhang2018collaborative}
Weichen Zhang, Wanli Ouyang, Wen Li, and Dong Xu.
\newblock Collaborative and adversarial network for unsupervised domain
  adaptation.
\newblock In \emph{CVPR}, 2018.

\bibitem[Zhao et~al.(2018)Zhao, Zhang, Wu, Moura, Costeira, and
  Gordon]{zhao2018adversarial}
Han Zhao, Shanghang Zhang, Guanhang Wu, Jos{\'e}~MF Moura, Joao~P Costeira, and
  Geoffrey~J Gordon.
\newblock Adversarial multiple source domain adaptation.
\newblock In \emph{NeurIPS}, 2018.

\bibitem[Zhao et~al.(2020)Zhao, Wang, Zhang, Gu, Li, Song, Xu, Hu, Chai, and
  Keutzer]{zhao2020multi}
Sicheng Zhao, Guangzhi Wang, Shanghang Zhang, Yang Gu, Yaxian Li, Zhichao Song,
  Pengfei Xu, Runbo Hu, Hua Chai, and Kurt Keutzer.
\newblock Multi-source distilling domain adaptation.
\newblock In \emph{AAAI}, 2020.

\bibitem[Zhou et~al.(2020)Zhou, Yang, Qiao, and Xiang]{zhou2020domain}
Kaiyang Zhou, Yongxin Yang, Yu~Qiao, and Tao Xiang.
\newblock Domain adaptive ensemble learning.
\newblock \emph{arXiv preprint arXiv:2003.07325}, 2020.

\bibitem[Zhou et~al.(2021)Zhou, Liu, Qiao, Xiang, and Loy]{zhou2021domain}
Kaiyang Zhou, Ziwei Liu, Yu~Qiao, Tao Xiang, and Chen~Change Loy.
\newblock Domain generalization: A survey.
\newblock \emph{arXiv preprint arXiv:2103.02503}, 2021.

\bibitem[Zhu et~al.(2017)Zhu, Park, Isola, and Efros]{zhu2017unpaired}
Jun-Yan Zhu, Taesung Park, Phillip Isola, and Alexei~A Efros.
\newblock Unpaired image-to-image translation using cycle-consistent
  adversarial networks.
\newblock In \emph{ICCV}, 2017.

\bibitem[Zhu et~al.(2019)Zhu, Zhuang, and Wang]{zhu2019aligning}
Yongchun Zhu, Fuzhen Zhuang, and Deqing Wang.
\newblock Aligning domain-specific distribution and classifier for cross-domain
  classification from multiple sources.
\newblock In \emph{AAAI}, 2019.

\bibitem[Zou et~al.(2019)Zou, Yu, Liu, Kumar, and Wang]{zou2019confidence}
Yang Zou, Zhiding Yu, Xiaofeng Liu, BVK Kumar, and Jinsong Wang.
\newblock Confidence regularized self-training.
\newblock In \emph{ICCV}, 2019.

\end{thebibliography}

\newpage

\appendix
\appendixpage

\renewcommand{\appendixname}{~\Alph{section}}
\setcounter{table}{0}
\setcounter{figure}{0}
\setcounter{equation}{0}
\renewcommand{\thetable}{\Alph{table}}
\renewcommand{\thefigure}{\Alph{figure}}
\renewcommand{\theequation}{\Alph{equation}}

The supplementary material is organized as follows: Section~\ref{sec:training_detail} gives the experimental setting of our experiments; Section~\ref{sec:tcl_src} evaluates the class compactness loss; 
Section~\ref{sec:dac_vs_fda} provides more experiments and analyses for the comparison between domain attention consistency (DAC) and feature distribution alignment (FDA); Section~\ref{sec:single_adapt} evaluates our DAC-Net on single-source to single-target domain adaptation task; Section~\ref{sec:src_domain_num} investigates how the number of source domains influences the performance of DAC-Net; Section~\ref{sec:amount_target_data} further explores the influence of the amount of target domain data on our DAC-Net; Section~\ref{sec:attend_feat_dist} shows visualizations on feature distribution.

\section{Experimental Setting}\label{sec:training_detail}

\begin{figure}[t]
    \centering
    \includegraphics[width=0.7\columnwidth]{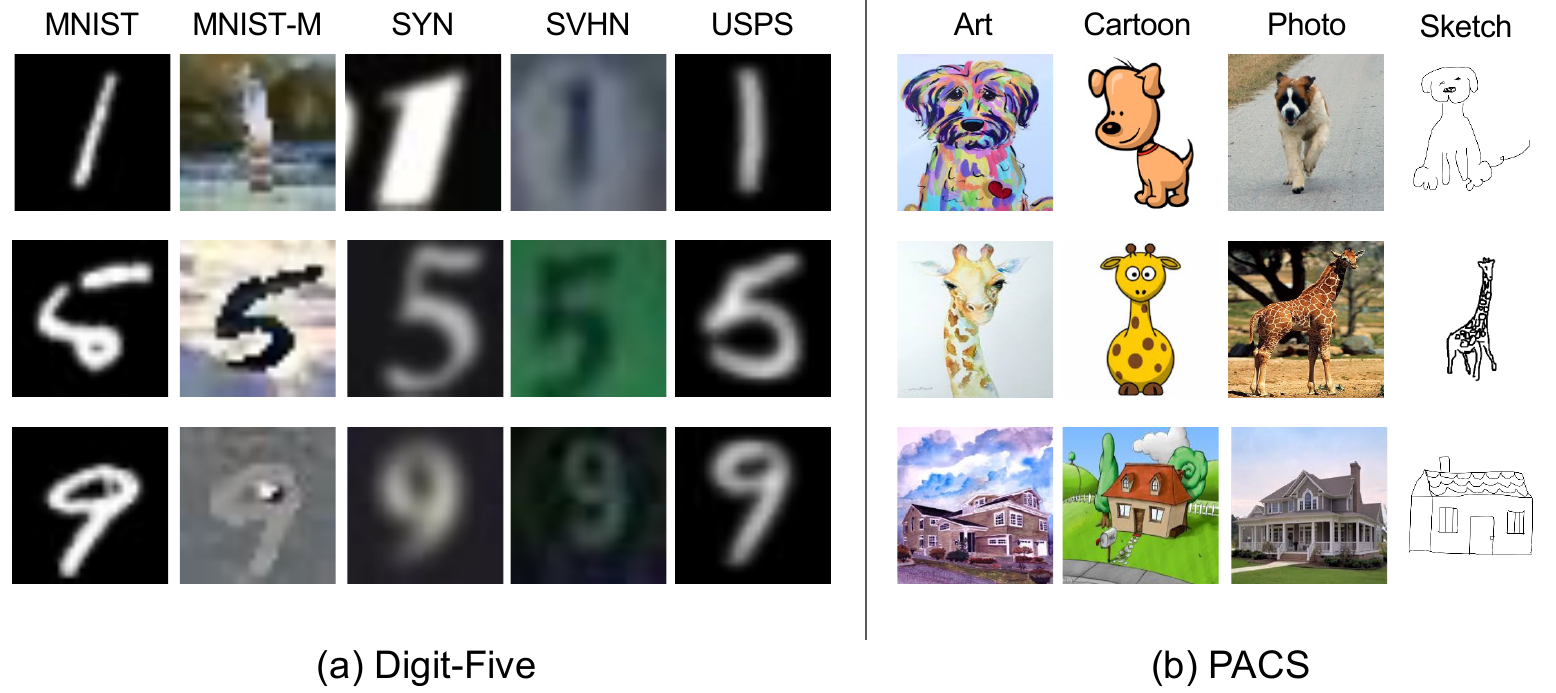}
    \vspace{-0.1cm}
    \caption{Example images from Digit-Five and PACS.}
    \label{fig:ims_digit_pacs}
\end{figure}

\paragraph{Datasets and protocols.}
(1) \textbf{DomainNet} is by far the largest and most challenging MSDA dataset. It has around 0.6 million images and 345 categories. There are six distinct domains (Clipart, Infograph, Painting, Quickdraw, Real and Sketch) with dramatic domain shift in image style, color, background, strokes, etc. See Figure 1 in the main paper for example images. It can be observed that some domains, e.g., Quickdraw and Sketch where shape information is particularly important, are closer than others like Real and Quickdraw where color/texture information is essential for the former but not required at all for the latter. Therefore, to succeed in DomainNet one needs to identify features that are transferable between the target and  source domains.
(2) \textbf{Digit-Five} consists of five digit datasets including MNIST~\cite{1998Gradient}, MNIST-M~\cite{syn_digits}, Synthetic Digits~\cite{syn_digits}, SVHN~\cite{37648}, and USPS. Following~\cite{Peng_2019_ICCV}, all 9,298 images in USPS are used as source domain for model training only; on each of the other four datasets, 25,000 images are used for training and 9,000 images for testing. The domain shift mainly takes place in font color/style and image background. See Figure~\ref{fig:ims_digit_pacs}(a) for example images.
(3) \textbf{PACS}~\cite{li2017deeper} includes 9,991 images of seven categories across four domains, i.e.~Cartoon, Photo, Sketch and Art Painting. The domain shift mainly corresponds to image style changes. Example images are provided in Figure~\ref{fig:ims_digit_pacs}(b). We follow the official train-val splits provided in~\cite{li2017deeper}.

For evaluation, one domain is chosen in turn as the target domain while the rest are regarded as source domains. A model is trained using the labeled source data and the unlabeled target data training split, and  tested on the test split of the target domain.

\paragraph{Training details.}
On Digit-Five and DomainNet, we use SGD with momentum to train the model. The learning rate is updated with the cosine annealing strategy~\cite{loshchilov2016sgdr}; on PACS, we use Adam~\cite{kingma2014adam} as the optimizer.

On DomainNet, ResNet-101~\cite{7780459} is used as the CNN backbone, following~\cite{Peng_2019_ICCV}. We sample 6 images from each domain at each iteration. The model is trained for 40 epochs, with an initial learning rate of 0.002.


On Digit-Five, we follow~\cite{Peng_2019_ICCV} to construct the CNN model with three convolution layers followed by two fully connected layers. The batch size $B$ is set to 64 for each domain. We train the model for 30 epochs, with an initial learning rate of 0.05.

On PACS, we follow the setting in~\cite{wang2020learning} and use ResNet-18~\cite{7780459} as the CNN backbone. The model is trained for 100 epochs with an initial learning rate of 5e-4 and the batch size of 16.

\section{Ablation Study on class compactness Loss}
\label{sec:tcl_src}

In this section, we first compare our class compactness loss with other related work, and then discuss whether this loss should be applied to source domain data.

JDDA~\cite{chen2019joint} and HoMM~\cite{chen2020homm} are the most related works to our class compactness loss. Both of them use parameterized class centers for discriminative feature learning. Instead, our class compactness loss takes advantage of classification weight vectors, which shows superior performance to class-center-based methods (see Table 2 in the main paper).

In practice, we can apply $L_c$ to both source and target domains and denote it as $L'_c$, i.e.
\begin{equation}
\begin{aligned} \label{eq:Lc_src}
L'_c =  \frac{1}{B} \sum_{i=1}^B \mathbbm{1}(q(\hat{y}_i^{\mathcal{T}}) \geq \tau) ||f_i^{\mathcal{T}} - W_{\hat{y}_i^{\mathcal{T}}} ||_2^2 +
 \frac{1}{KB} \sum_{k=1}^K\sum_{i=1}^B  ||f_i^{\mathcal{S}_k} - W_{y_i^{\mathcal{S}_k}} ||_2^2.
\end{aligned}
\end{equation}
We then compare the $L_c$  and $L'_c$ in Table~\ref{tab:ablation_cl}. We can see that applying the compactness loss to source domains decreases the accuracy by 2.03\%. The degradation implies that the compactness for source domain data could even harm the learning of the discriminative features for the target data.

Furthermore, we compare our class compactness loss with another loss which is a $\ell_{2}$ penalty on the weight matrix $W_{y_i}$. We can see from Table~\ref{tab:ablation_cl} that such a loss works worse than our class compactness loss. Our class compactness works better because it pulls the features of target data to the corresponding classification weight vectors, thus possibly away from the decision boundary.

\begin{table}[t]
    \centering
    \begin{tabular}{l|c}
    \hline
      Methods & \textbf{Avg}\\
    \hline
       $L_c$ (only target data) & 91.42\\
       $L'_c$ (both source and target data) & 89.39\\
       $\ell_{2}$ penalty on the weight $W_{y_i}$ & 90.91\\
    \hline
    \end{tabular}
    \caption{Ablation study on whether applying the compactness loss to source domain data on PACS.}
    \label{tab:ablation_cl}
    \vspace{-0.2cm}
\end{table}

\section{Attention Alignment vs.~Feature Alignment}
\label{sec:dac_vs_fda}

\begin{table*}[t]
    \centering
    \begin{tabular}{l|l|c}
    \hline
      Methods & Apply domain alignment loss after & \textbf{Avg}\\
    \hline
        DAC & $L_d$ after last two residual blocks	&\textbf{90.79}\\
    \hline
        \multirow{8}{*}{FDA} & $L_d$ after final features & 89.27\\
        &  $L_d$ after last two residual blocks	&89.75\\
        &  $L_d$ after last three residual blocks & 88.75\\
        &  $L_d$ after all four residual blocks & 88.78\\
        & MMD loss after final features & 89.65\\
        & MMD loss after last two residual blocks	& 89.23\\
        & MMD loss after last three residual blocks	& 89.80\\
        & MMD loss after all four residual blocks	& 89.74\\
    \hline
    \end{tabular}
    \caption{Ablation study on where to apply the domain alignment loss in the ResNet architecture on PACS. FDA: Feature Distribution Alignment.}
    \label{tab:ablation_fda}
\end{table*}

In Table~\ref{tab:ablation_fda}, we compare our DAC loss with different variants of the FDA loss. For a fair comparison, the experimental setup for all methods is kept the same. For the FDA-based methods, we impose the domain alignment loss on the average-pooled features to directly align feature distributions. Two different domain alignment losses are explored here. The first is $L_d$ (defined in Eq.~(2)), which is applied to features instead of attribute attention weights. The second is maximum mean discrepancy (MMD)~\cite{gretton2012kernel}. From the results, we can see that DAC surpasses all FDA variants by about 1\%. This suggests that aligning attribute attention weights is more effective than aligning feature distributions for MSDA.

\section{Single-Source to Single-Target Domain Adaptation}\label{sec:single_adapt}
We further evaluate our DAC on popular single-source domain adaptation datasets, namely Digits datasets. Following the setting in ~\cite{xiao2021dynamic}, we conduct experiments on three transfer tasks: SVHN $\rightarrow$ MNIST, USPS $\rightarrow$ MNIST, MNIST $\rightarrow$ USPS. As in~\cite{2018Maximum, xiao2021dynamic}, a three-layer CNN is used for the first task while a two-layer CNN for the latter two tasks.
We then apply channel attention to these CNNs to facilitate DAC. Other training details are kept the same as ~\cite{2018Maximum,xiao2021dynamic}: Adam~\citep{kingma2014adam} optimizer with learning rate of 2e-4, mini-batch size of 128. The results are shown in Table~\ref{tab:single_src}. We can see that our DAC effectively improves the accuracy by 5.4\% over the model without DAC.

In Table~\ref{tab:pacs_single_best}, we also compare the MSDA task with other single-source domain adaptation tasks on PACS, including single best and source combine. Single best is the best results of adapting each of three source domains to the target, and source combine means combining all source domains as a single source domain. It is obvious that the our DAC-Net works best under the setting of MSDA. This is because the transferable attributes learning of our DAC-Net can effectively alleviate the negative transfer, and take full advantage of multiple source domains to improve the adaptation performance.

\begin{table}[t]
    \centering
    \begin{tabular}{l|p{1.5cm}p{1.5cm}p{1.5cm}|c}
    \hline
      \textbf{Methods} & \textbf{SVHN$\rightarrow$ MNIST} &\textbf{USPS$\rightarrow$ MNIST} & \textbf{MNIST$\rightarrow$ USPS} & \textbf{Avg}\\
    \hline
       w/ DAC & 77.82	&92.25	&89.91	&86.66\\
       w/o DAC & 68.79	&94.44	&80.55	&81.26\\
    \hline
    \end{tabular}
    \caption{Accuracy on Digits dataset for single-source domain adaption.}
    \label{tab:single_src}
\end{table}

\begin{table}[t]
    \centering
    \begin{tabular}{l|cccc|c}
    \hline
    \textbf{Methods} & \textbf{ArtPainting} & \textbf{Cartoon} & \textbf{Sketch} & \textbf{Photo} & \textbf{Avg}\\
    \hline \hline
    DAC-Net (Single best) &86.67 &78.00 &68.30 &97.54 & 82.63\\
    DAC-Net (Source combined) &\textbf{91.62} &89.53 &82.07 &\textbf{98.48} &90.43\\
    DAC-Net (MSDA) &91.39	&\textbf{91.39}	&\textbf{84.97}	&97.93	&\textbf{91.42} \\
    \hline
    Oracle &99.53 &99.84 &99.53 &99.92 & 99.71\\
    \hline
    \end{tabular}
    \caption{Accuracy on PACS. Best results for each target domain (except oracle) are in bold.}
    \label{tab:pacs_single_best}
\end{table}

\section{Influence of the Number of Source Domains}\label{sec:src_domain_num}
In this section, we gradually increase the number of source domains to see how it influences the accuracy of our DAC-Net. We show the results in Table~\ref{tab:pacs_num_src_domain}. The findings can be summarized as follows: (1) Generally, more source domains increase the performance on the target, possibly because more source domains contribute to the learning of transferable semantic attributes. (2) When the target domain is Photo, the Sketch domain brings a minor negative impact on the performance (-0.09\%, i.e. 98.02\% vs. 97.93\%).  This may be caused by that the attributes learned from Sketch domain are too abstract to transfer to the concrete attributes of Photo. Overall, our DAC-Net can benefit from more source domains.

\begin{table}[tb]
    \centering
    \tabstyle{4pt}
    \begin{tabular}{cc|cc|cc|cc}
    \hline
    \textbf{Tasks} & \textbf{Avg} & \textbf{Tasks} & \textbf{Avg} & \textbf{Tasks} & \textbf{Avg}& \textbf{Tasks} & \textbf{Avg}\\
    \hline \hline
    \textbf{C$\rightarrow$A} &87.55 & \textbf{A$\rightarrow$C} &78.00 & \textbf{C$\rightarrow$S} &68.30 &\textbf{C$\rightarrow$P} &94.19 \\
    \textbf{C+S$\rightarrow$A} &88.67 & \textbf{A+S$\rightarrow$C} &90.19 & \textbf{C+A$\rightarrow$S} &78.88 &\textbf{C+A$\rightarrow$P} &98.02 \\
    \textbf{C+S+P$\rightarrow$A} &91.39 & \textbf{A+S+P$\rightarrow$C} &91.39 & \textbf{C+A+P$\rightarrow$S} &84.97 &\textbf{C+A+S$\rightarrow$P} &97.93 \\
    \hline
    \end{tabular}
    \caption{Ablation study on how the number of source domains influence the accuracy of DAC-Net on PACS. `C': Cartoon, `A': ArtPainting, `S': Sketch, `P': Photo.}
    \label{tab:pacs_num_src_domain}
\end{table}

\begin{table}[htb]
    \centering
    \tabstyle{3pt}
    \begin{tabular}{c|cccc|c}
    \hline
    \textbf{Amount of Target Data} & \textbf{ArtPainting} & \textbf{Cartoon} & \textbf{Sketch} & \textbf{Photo} & \textbf{Avg}\\
    \hline \hline
    100\% &91.39	&\textbf{91.39}	&\textbf{84.97}	&97.93	&\textbf{91.42} \\
    70\% &\textbf{92.11} &90.46 &80.36 &97.69 &90.16 \\
    50\% &91.26 &90.57 &76.78 &\textbf{98.22} &89.21 \\
    30\% &91.04 &88.72 &80.05 &97.54 &89.34\\
    10\% &86.34 &85.57 &77.25 &97.43 &86.65\\
    \hline
    \end{tabular}
    \caption{Accuracy of DAC-Net on PACS. }
    \label{tab:pacs_tar_amount}
\end{table}

\begin{figure}[htb]
    \centering
    \subfigure[DAC-Net]{
    \includegraphics[trim=60 60 60 60,clip,width=0.25\textwidth]{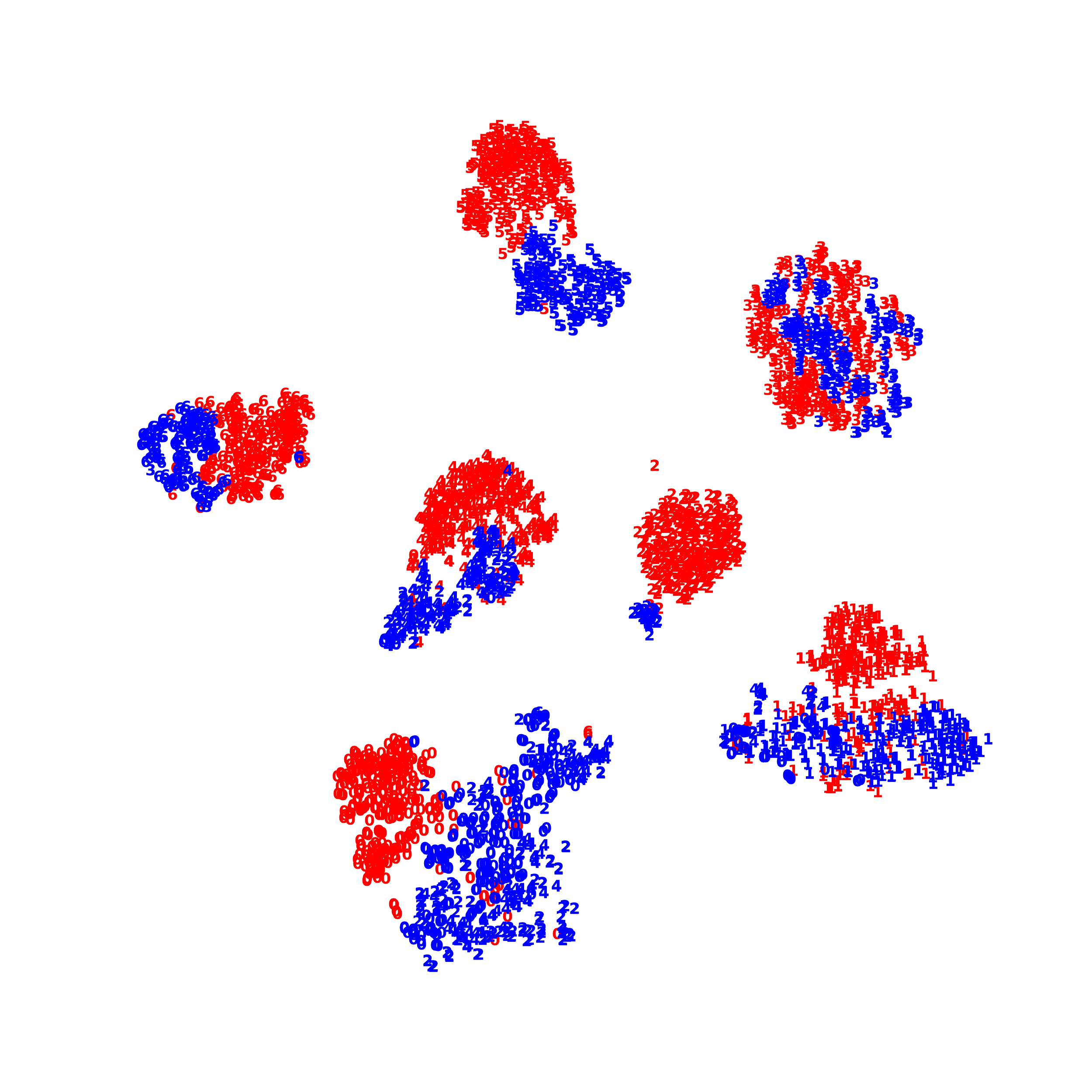}
    }
    \hspace{0.15cm}
    \subfigure[M$^3$SDA]{
    \includegraphics[trim=60 60 60 60,clip,width=0.25\textwidth]{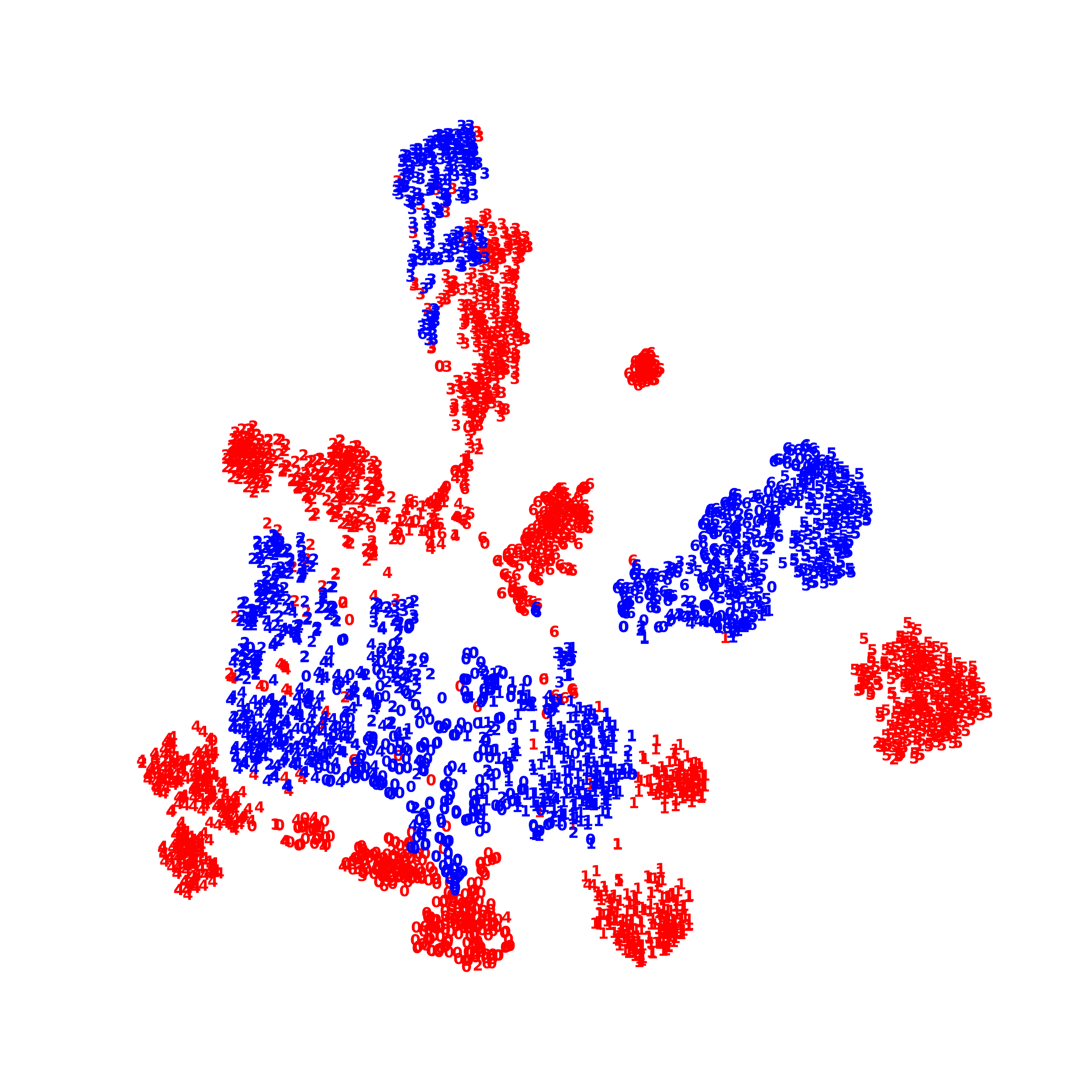}
    }
    \hspace{0.15cm}
    \subfigure[MCD]{
    \includegraphics[trim=60 60 60 60,clip,width=0.25\textwidth]{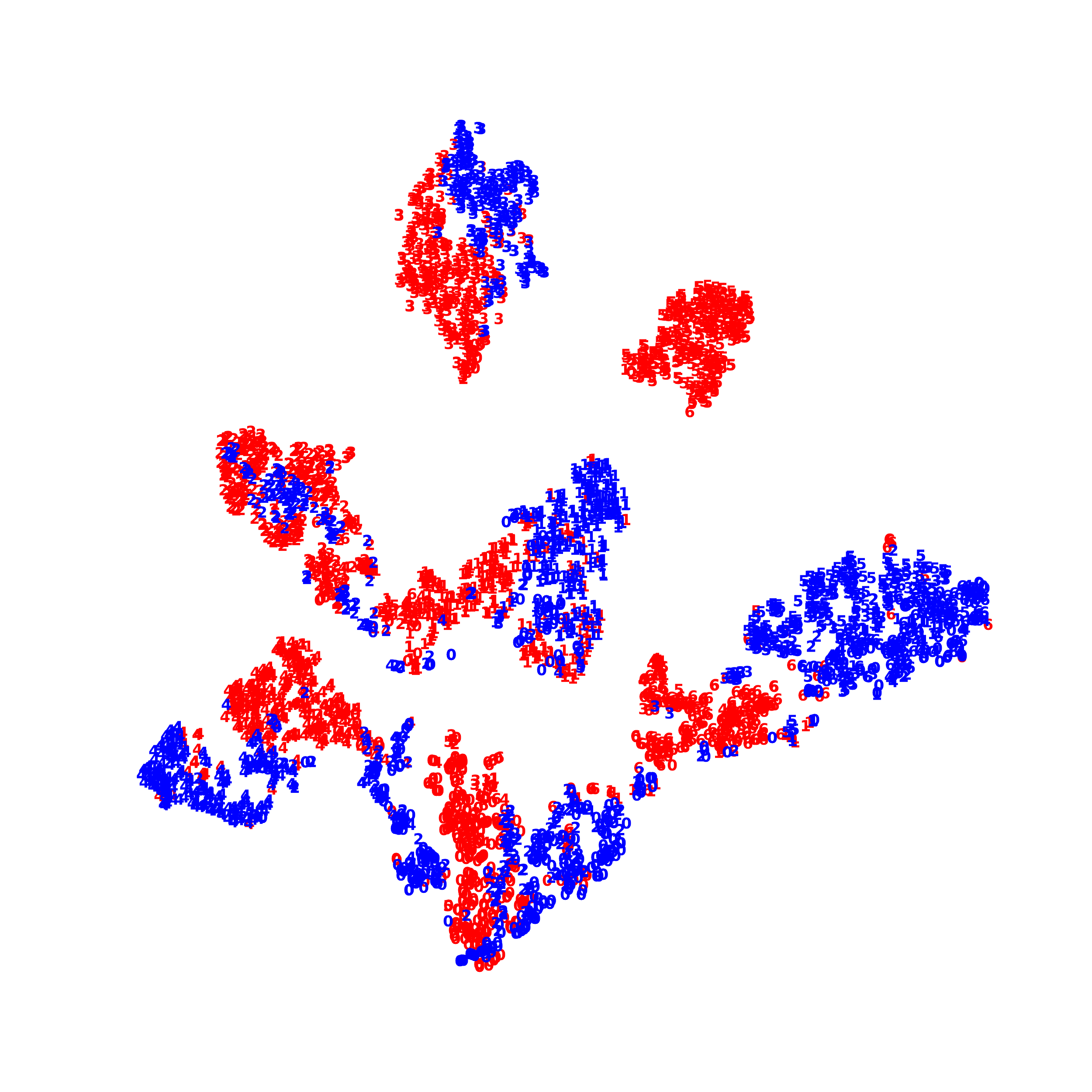}
    }
    \caption{Visualization of feature distributions of different methods on PACS using t-SNE~\cite{maaten2008visualizing}. Each number stands for the class label of a feature (7 classes in total; better viewed with zoom-in). Red/blue denotes the source/target domain.}
    \vspace{-0.2cm}
    \label{fig:visual_feat_dist}
\end{figure}

\section{Influence of the Amount of Target Domain Data}\label{sec:amount_target_data}
To investigate how many target data are required for our DAC-Net (under MSDA setting), we gradually reduce the amount of target domain data. From Table~\ref{tab:pacs_tar_amount}, we can see that the less target data leads to worse performance. But overall, our DAC-Net can beat the source-only baseline even when only 10\% amount of target data are used, e.g. 86.65\% vs. 81.94\%.

\section{Visualization}\label{sec:attend_feat_dist}

In this section, we provide visualizations of feature distribution to help understand why our DAC-Net works.

Figure~\ref{fig:visual_feat_dist} depicts the feature distributions of our DAC-Net,  M$^3$SDA~\cite{Peng_2019_ICCV} and  MCD~\cite{2018Maximum}.
MCD performs class-level feature distribution alignment across domains while M$^3$SDA only does alignment at the domain level. Figure~\ref{fig:visual_feat_dist} shows that both of them cannot achieve either class-level or domain-level feature alignment. As a result, the target domain data of 7 classes are not well separable. In contrast, our DAC-Net does not enforce any feature alignment, but the source and target features are much better aligned. More importantly, the 7 classes of all domains are easily separable using the same set of decision boundaries---after all, all domains share the classifier. This suggests that: (1) explicitly enforcing feature alignment is counter-productive since it even \emph{harms} the discriminative feature learning for classification task; (2) learning latent attributes by enforcing attention weight consistency is more effective for MSDA because it properly reduces the domain shift, which further \emph{contributes to} discriminative features learning on the target domain.


\end{document}